\documentclass[sigconf]{acmart}
\usepackage{CJK} % chinese
\usepackage{algorithm}
\usepackage{algorithmic}
\usepackage{natbib}

\usepackage{amsmath}
\usepackage{graphics}
\usepackage{epsfig}
\usepackage{multirow}
\usepackage{xspace}
\usepackage{subfigure}
\usepackage{fancyhdr}

\AtBeginDocument{%
  \providecommand\BibTeX{{%
    \normalfont B\kern-0.5em{\scshape i\kern-0.25em b}\kern-0.8em\TeX}}}

\copyrightyear{2021}
\acmYear{2021}
\setcopyright{acmcopyright}\acmConference[WSDM '21]{Proceedings of the Fourteenth ACM International Conference on Web Search and Data Mining}{March 8--12, 2021}{Virtual Event, Israel}
\acmBooktitle{Proceedings of the Fourteenth ACM International Conference on Web Search and Data Mining (WSDM '21), March 8--12, 2021, Virtual Event, Israel}
\acmPrice{15.00}
\acmDOI{10.1145/3437963.3441819}
\acmISBN{978-1-4503-8297-7/21/03}

\newcommand{\model}{TRIDENT\xspace}

\begin{document}
\fancyhead{}
\begin{CJK*}{UTF8}{gbsn}
\title[Triangular Bidword Generation for Sponsored Search Auction]{Triangular Bidword Generation for Sponsored\\ Search Auction}
% \title{Triangular Bidword Generation for \\Sponsored Search Auction}

\author{Zhenqiao Song, Jiaze Chen, Hao Zhou, Lei Li}
\affiliation{%
  \institution{ByteDance AI Lab}
%   \streetaddress{No.1999, Yishan Road}
%   \city{Shanghai}
%   \state{China}
%   \postcode{201203}
}
\email{{songzhenqiao,chenjiaze,zhouhao.nlp,lileilab}@bytedance.com}
% \author{
%   Zhenqiao Song, Jiaze Chen, Hao Zhou, Lei Li \\
%   ByteDance AI Lab \\
%   {\tt \{songzhenqiao, chenjiaze, zhouhao.nlp, lilei.02\}@fudan.edu.cn} }

% \renewcommand{\shortauthors}{Zhenqiao Song, Jiaze Chen, Hao Zhou, Lei Li}
%\renewcommand{\shortauthors}{Zhenqiao Song, Jiaze Chen, Hao Zhou, Lei Li}

\begin{abstract}
Sponsored search auction is a crucial component of modern search engines. It requires a set of candidate bidwords that advertisers can place bids on.
Existing methods generate bidwords from search queries or advertisement content. However, they suffer from the data noise in <query, bidword> and <advertisement, bidword> pairs.
In this paper, we propose a triangular bidword generation model (TRIDENT), which takes the high-quality data of paired <query, advertisement> as a supervision signal to indirectly guide the bidword generation process.
Our proposed model is simple yet effective: by using bidword as the bridge between search query and advertisement, the generation of search query, advertisement and bidword can be jointly learned in the triangular training framework. 
This alleviates the problem that the training data of bidword may be noisy.
Experimental results, including automatic and human evaluations, show that our proposed \model\ can generate relevant and diverse bidwords for both search queries and advertisements.
Our evaluation on online real data validates the effectiveness of the \model's generated bidwords for product search.
\end{abstract}




\keywords{sponsored search, advertising bidword generation, query expansion, triangle  training}


\maketitle

\section{Introduction}
\label{sec:background}

\begin{figure}[t]
  \centering
  \includegraphics[width=7.5cm]{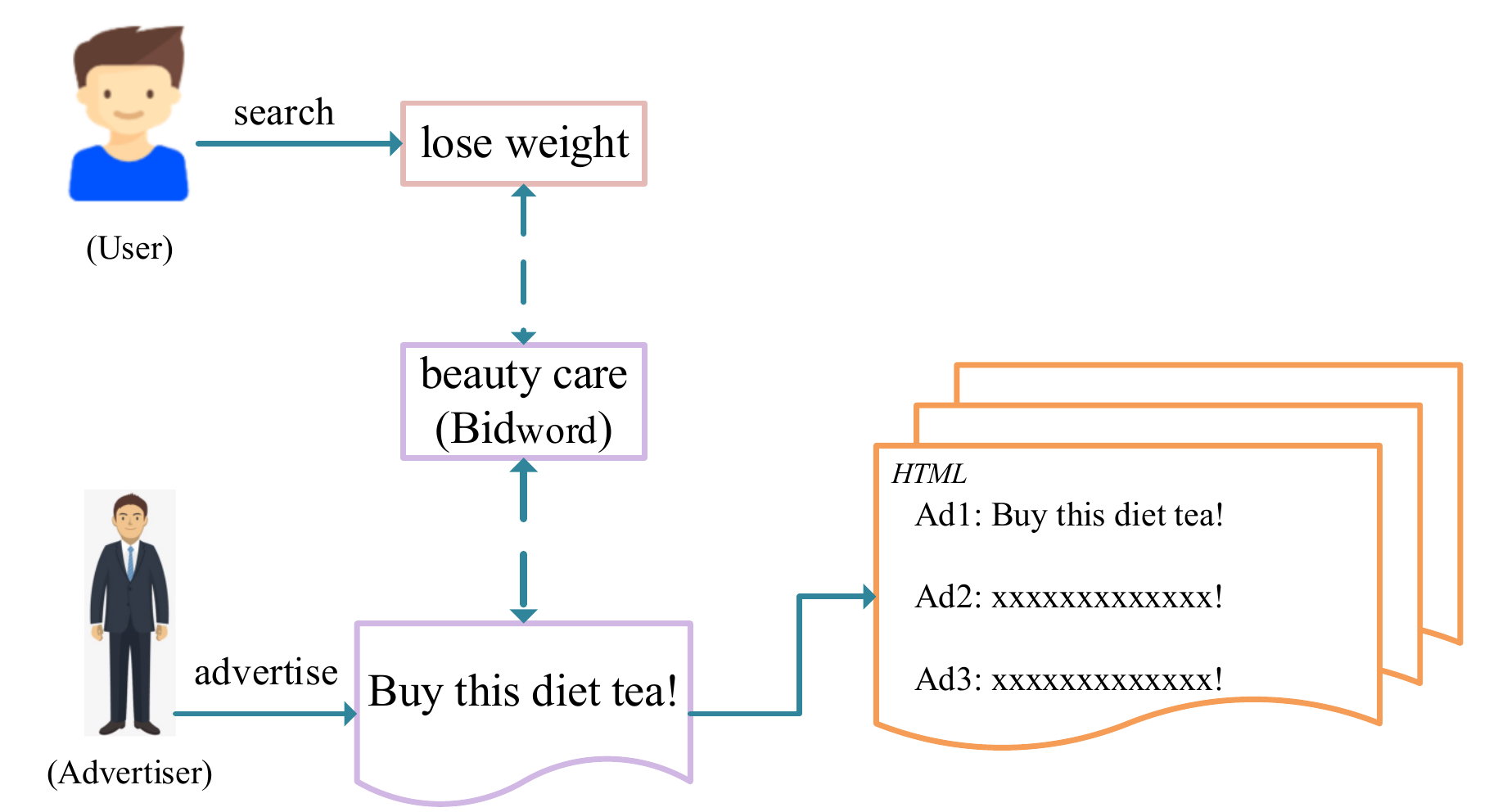}
  \caption{An illustration of sponsored search auction system. Here ``lose weight'' is the search query from user, and ``beauty care'' is the bidword purchased by advertiser.}
  \label{figure1}
\end{figure}

Sponsored search auction is an indispensable part of commercial search engine, which aims to recommend appropriate advertisements to search users.
Sponsored search auction is also called bidword auction, as it takes \textit{bidwords} as bridges between search query and advertisements~\cite{aggarwal2006truthful,lian2019end}, which is widely adopted in online advertising by search companies.
Specifically, a search engine first retrieves a set of advertisements whose associated bidwords match a user issued query.
It then ranks these advertising candidates according to an auction process by considering both the quality and the bid price of each advertisement~\cite{aggarwal2006truthful,lian2019end}.
Finally, the top-ranked advertisements will be presented in the search page.
For example, in Figure \ref{figure1}, advertisers bid a so-called bidword for their product in the auction system, and the so-called bidword is also associated with a search query according to some match methods.
Thus, when a user delivers this query in the search engine, the advertisement of this product will be triggered and inserted into the result search page.
Obviously, the so-called bidword plays an important role in sponsored search for connecting advertising~(Ad) and query.
Therefore, generating relevant and diverse bidwords for Ad and query are crucial, as it can improve the effectiveness and efficiency of advertising-bidword auction and query-bidword matching.
Through this way, high-quality bidwords can bring great revenues for sponsored search~\cite{joshi2006keyword,abhishek2007keyword,zhou2019domain,lian2019end,lee2018rare}.

% Generating relevant and diverse keywords for Advertisements~(Ad for short) is crucial for sponsored search.
% Generally, sponsored search is an interplay of three parts~\cite{lian2019end} as shown in Figure~\ref{figure1}.
% First, a search engine retrieves a set of advertisements whose keywords match a user submitted query.
% It ranks these advertising candidates with an auction process by considering both the advertising quality and the bidding price of each ad~\cite{aggarwal2006truthful}.
% Finally, the chosen advertisement is presented in a result page provided by the search engine.
% Therefore, high-quality ad and query keyword recommendation has the big potentiality to bring huge revenue to search companies.
%%再改改intro
%第一段说清楚了是什么，为啥两条线重要。
%现列举困难，有噪声的语料；
%然后说传统方法，首先传统的方法对稀疏数据不友好、不灵活无法生成新的bidword、不易扩展到大规模语料上；%其次基于深度学习的，适用于大规模语料，但现存的大规模语料质量不高，由此自然而然提出triangular framework

Generating satisfactory bidwords is non-trivial for sponsored search.
Traditional methods are mainly matching-based, which match the co-occurrence words between existed bidwords~\cite{chen2008advertising} and search query~\cite{fuxman2008using} (or advertisement~\cite{chen2008advertising}).
Different matching strategies have been used, including exact match~\cite{iacus2012causal}, broad match~\cite{even2009bid} and phrase match~\cite{liu2018domain,garber2004flexible}.
However, bidwords of Ads or queries belonging to a rare domain are hard to obtain using matching strategies, because there are barely overlapped words between these Ads (or queries) and existing bidwords.
Additionally, matching-based methods cannot recommend novel bidwords that do not exist in the search history.

Recently, deep learning approaches give effective alternatives for bidword generation, which could address the concerns of matching-based methods.
Given paired data such as <query, bidword> and <advertisement, bidword>, we can employ the \textit{sequence to sequence} model with \textit{encoder-decoder} framework from neural machine translation~\cite{sutskever2014sequence,cho2014learning,bahdanau2014neural} to directly generate bidwords from search queries or advertisements~\cite{grbovic2015context}.
The sequence to sequence~(Seq2Seq) model has been widely used and obtains gains in applications of modern search engine, such as advertising keyword suggestion~\cite{grbovic2015context}, query keyword generation~\cite{lian2019end} and query expansion~\cite{lee2018rare}.

\begin{figure}
  \centering
  \includegraphics[width=6.0cm]{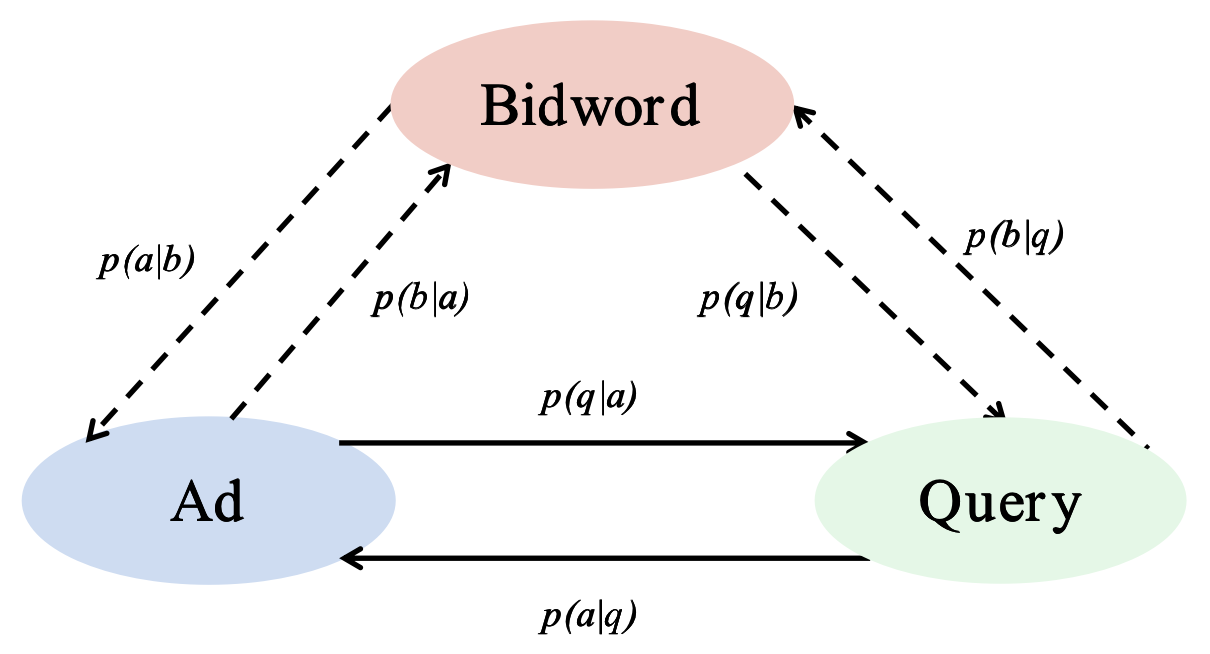}
  \caption{The architecture of the triangular bidword generation model (\model). The solid lines mean the data of the direction is extensive and high-quality, while the dash lines mean the data of the direction is noisy. $a$, $q$ and $b$ denote Ad, query and bidword, respectively.}
  \label{figure2}
\end{figure}

% However, current approaches of applying Seq2Seq model in Ad keywords generation do not give enough accuracy gains, due to that the training data maybe noisy and insufficient.
% Specifically, high quality training data of paired (query, keywords) or (advertisements, keywords) are hard to obtain, which is different to the case of machine translation~(more than 10 millions for high resource languages).
% High quality (query, keywords) or (advertisements, keywords) needs heavy human annotation, and noisy training data may prevent Seq2Seq model generating good Ad keywords for advertisements and queries.
% Fortunately, the paired data of <query, advertisement> are quite easy to obtain from the user-clicks of search engine, which is always high-quality. 
% How to fully exploit (query, advertisement) for Ad keyword generation is still under explored. 
However, current approaches of directly applying Seq2Seq in bidword generation do not achieve enough gains, due to the fact that the training data may be noisy.
Specifically, high-quality training data of paired <query, bidword> or <advertisement, bidword> are hard to obtain, which is caused by the common \textit{keyword bidding problem}\footnote{Keyword bidding problem: Advertisers are preferable to bid on more popular bidwords to improve the trigger probability of their Ads even if those bidwords are not relevant with their products.}.
Thus, noises are introduced into <query, bidword> or <advertisement, bidword> data, and noisy training data may prevent Seq2Seq from generating high-quality bidwords.
Fortunately, the paired data of <query, advertisement> are quite easy to obtain from the user-clicks of search engine, which are always high-quality. 
How to fully exploit <query, advertisement> data for bidword generation is still under explored.

In this paper, we propose a TRIangular biDword gENeratTion model~(\model), which can effectively leverage the high-quality data of paired <query, advertisement> to generate more relevant and diverse bidwords for both advertisements and queries.
Motivated by the research of low-resource neural machine translation~\cite{firat2016multi,firat2017multi,dong2015multi,ren2018triangular}, our proposed \model\ is simple yet effective, the intuition of which is very straightforward:
$P$(query | Ad) can be well learned by high-quality <query, advertisement> data, and they could also be estimated as Figure \ref{figure2} using the Maximum Marginal Likelihood~\cite{kovcisky2016semantic}:
$ \tilde{P}(\text{query}| \text{Ad})$ = $\sum_{\text{bidword}}$ $P(\text{bidword} | \text{Ad})$ $P(\text{query} | \text{bidword}) $.
Finally, $P$(bidword|Ad) can be indirectly supervised in the triangle by minimizing the divergence between $P(\text{query} | \\\text{Ad})$ and $\tilde{P}(\text{query}| \text{Ad})$.
$P(\text{bidword}| \text{query})$ could be learned in the same way.
Thus, information of high-quality <query, advertisement> data can also be used to generate bidwords, alleviating the problem that <query, bidword> and <advertisement, bidword> may be noisy.

Experimental results with both automatic and human evaluations show that bidwords generated by our TRIDENT are significantly better than baselines in both relevance and diversity.
The model performance can be further improved by using the proposed constrained beam search.
% \model has been deployed as a HTTP service for the online advertising at ByteDance.

Our contributions are listed as follows:
\begin{itemize}
    \item We propose \model, a novel triangular indirectly supervised bidword generation model, which can effectively boost the performance of bidword generation by exploiting high-quality <query, advertisement> data as an indirect supervision signal.
    \item Experimental results with both automatic and human evaluations demonstrate that our model can generate relevant and diverse bidwords for both advertisements and queries.
    \item To our best knowledge, TRIDENT is the first work to employ a triangular training framework for bidword generation.
\end{itemize}

\section{Background}
\label{sec:background}
\subsection{Notation}
In the following formulations, we use $A$, $Q$ and $B$ to denote Ad, query and bidword, respectively.
Suppose <$a, q$> is a pair of high-quality <Ad, query> data, we denote the <Ad, query> dataset as $D_{aq}=\{\text{<}a,q\text{>}_i\}_{i=i}^N$ and $N$ is size of $D_{aq}$.
Likewise, $D_{ab}=\{\text{<}a,b\text{>}_i\}_{i=1}^M$ and $D_{qb}=\{\text{<}q,b\text{>}_i\}_{i=1}^L$ can be denoted for the <Ad, bidword> and <query, bidword> datasets in the same way.

\subsection{Encoder-Decoder Framework}
The encoder-decoder framework is a widely used generative neural network architecture, which is first introduced in Neural Machine Translation~\cite{sutskever2014sequence,cho2014learning}.
Here we implement our method based on Transformer~\cite{vaswani2017attention} encoder-decoder framework.
The encoder first transforms the input sequence $X=\{x_1, x_2, ..., x_n\}$ into a sequence of feature vectors $Z=\{z_1, z_2, ..., z_n\}$, from which the decoder generates an output sequence $Y=\{y_1, y_2, ..., y_m\}$. 
The encoder and decoder are trained jointly to maximize the conditional probability of $Y$ given $X$:
\begin{equation}
\small
P(Y|X)=\sum _{j=1}^m\log P(y_j|y_{<j}, X;\theta)
\label{equation23}
\end{equation}

Specifically, Transformer consists of stacked encoder and decoder layers.
The encoder layer is a self-attention block followed by a position-wise feed-forward block, based on which the decoder layer has an extra encoder-decoder attention block.
For self-attention and encoder-decoder attention, a multi-head attention block is used to obtain information from different representation subspaces at different positions.
Each head corresponds to a scaled dot-product attention, which operates on query \textit{Q}, key \textit{K} and value \textit{V}:
\begin{equation}
\small
{\rm Attention}(Q, K, V) = {\rm softmax}(\frac{Q K^T}{\sqrt{d_k}})V
\label{equation24}
\end{equation}
where $d_k$ is the dimension of the key.
Then the heads are concatenated and once again projected, resulting in the final values, as described in the following formulation:
\begin{equation}
\small
\begin{split}
  {\rm MultiHead}(Q, K, V) &= {\rm Concat(head_1, ..., head_h)}W^O  \\
  {\rm head_i} &= {\rm Attention}(Q W_i^Q, K W_i^K, V W_i^V)
\end{split}
\label{equation25}
\end{equation}
where $W_i^Q$, $W_i^K$, $W_i^V$ and $W^O$ are trainable parameters, and $h$ is the number of heads.

\section{Proposed Method: \model}
\label{sec:method}
In this section, we describe our proposed triangular bidword generation model TRIDENT, which is able to generate relevant and diverse bidwords for both Ads and queries. 
The overall architecture is shown in Figure\ref{figure2}.

Generally TRIDENT includes six generation processes.
It first starts with modeling the generation task $A\Rightarrow Q$ on the high-quality <$A$, $Q$> data, leading to a generation probability $P(q|a)$ for each <$a$, $q$> pair.
Then we decompose $A\Rightarrow Q$ into two phases by training another two generation models $A\Rightarrow B$ and $B\Rightarrow Q$ on the noisy data pairs (<$A$, $B$> and <$Q$, $B$>).
By taking $B$ as an intermediate latent variable, we can calculate another generation probability $\tilde{P}(q|a)$.
Then our model will be trained to eliminate the divergence between the direct probability $P(q|a)$ and the indirect one $\tilde{P}(q|a)$. 
Through this way, $P(b|a)$ can be indirectly supervised in the triangular architecture among $A$, $Q$ and $B$.
The situation is similar by reversing the direction of $A\Rightarrow Q$, and thus the quality of generated bidwords with noisy data can be improved.

In the following subsections, we will first introduce the training of the triangular architecture.
Then we describe how to construct bidword as an intermediate latent variable.
Next, we will discuss more training details and present our algorithm in the form of pseudo code.
Finally, a constrained beam search is proposed to further improve the diversity of the bidwords.

\subsection{Triangular Model Training}

Since it is difficult to obtain high-quality <$A$, $B$> and <$Q$, $B$> data, we propose a triangular architecture (TRIDENT) to promote bidword generation with the help of high-quality <$A$, $Q$> data.

Overall, there are six generation models among Ad, query and bidword in the proposed TRIDENT, as depicted in Figure\ref{figure2}.
These models all employ the Transformer encoder-decoder framework, and are jointly trained through the following training objective:
\begin{equation}
L = \lambda L_{MLE} + (1- \lambda) L_{TRI} 
\label{equation1}
\end{equation}
where $\lambda$ is a hyper-parameter to govern the relative importance of the two losses, $L_{MLE}$ is the likelihood loss of generation models on the noisy data (<$A$, $B$> and <$Q$, $B$>), and $L_{TRI}$ is the divergence between the direct generation probability on <$A$, $Q$> data and the indirect one computed by taking $B$ as an intermediate latent variable in the triangle.

% Specifically, assuming $a, b, q$ denote advertisements, bidwords and search query, respectively. 
Specifically, $L_{MLE}$ can be computed as:
\begin{equation}
\begin{split}
L_{MLE} = &-\sum_{<a,b>\in D_{ab}}[P(a|b;\theta_{ba}) + P(b|a;\theta_{ab})] \\
&- \sum_{<q,b>\in D_{qb}}[P(q|b;\theta_{bq}) + P(b|q;\theta_{qb})]
\end{split}
\end{equation}
where $P(b|a;\theta_{ab})$ is the conditional probability of a bidword $b$ given an Ad $a$, $\theta_{ab}$ is the corresponding parameters, and so on. 
All the probabilities are calculated using Equation\ref{equation23} with their corresponding parameters $\theta$.
Through $L_{TRI}$, we aim to minimize the cross-entropy of direct generation probability on <$A$, $Q$> and the indirect one:
\begin{equation}
\begin{split}
L_{TRI} = -\sum_{<a,q>\in D_{aq}}&\Big[P(q|a;\theta_{aq})\ \log \tilde{P}(q|a;\theta_{ab}, \theta_{bq}) \\
& + P(a|q;\theta_{qa})\ \log \tilde{P}(a|q;\theta_{qb}, \theta_{ba})\Big]
\end{split}
\label{equation555}
\end{equation}
Here $\tilde{P}(q|a;\theta_{ab}, \theta_{bq})$ and $\tilde{P}(a|q;\theta_{qb}, \theta_{ba})$ are the indirect probabilities between $A$ and $Q$, which are computed by taking $B$ as an intermediate variable:
\begin{equation}
\begin{split}
\tilde{P}(q|a;\theta_{ab}, \theta_{bq}) &= \sum_b P(q|b;\theta_{bq})P(b|a;\theta_{ab}) \\
\tilde{P}(a|q;\theta_{qb}, \theta_{ba}) &= \sum_b P(a|b;\theta_{ba})P(b|q;\theta_{qb})
\label{equation2}  
\end{split}
\end{equation}

In detail, TRIDENT first directly train two generation models between $A$ and $Q$, after which these two models are frozen and will  indirectly supervise the subsequent bidword generation process by regarding $B$ as a bridge connecting $A$ and $Q$.
Instead of training two separated generation models, we build the bi-directional dependencies between $A$ and $Q$ simultaneously.
Inspired by previous works on dialogue systems \cite{li2016diversity,shen2018nexus} , we propose to maximize the mutual information\footnote{Available at http://en.wikipedia.org/wiki/Mutual\_information/} between $A$ and $Q$ to model their bi-directional dependencies.
To simplify computation, we maximize the following lower bound of mutual information between $A$ and $Q$:
\begin{equation}
\begin{split}
I(A,Q) &= \frac{1}{2}[\sum_{<a,q>}P(a,q)\log\frac{P(a,q)}{P(a)}-\sum_{<a,q>}P(a,q)\log P(q)] \\
&+ \frac{1}{2}[\sum_{<a,q>}P(a,q)\log\frac{P(a,q)}{P(q)}-\sum_{<a,q>}P(a,q)\log P(a)] \\
&\geq \frac{1}{2}[\sum_{<a,q>}P(a,q)\log P(q|a) + \sum_{<a,q>}P(a,q)\log P(a|q)] \\
\end{split}
\label{equation3}
\end{equation} 
Suppose $A$ and $Q$ are both sampled from a uniform distribution and thus the above lower bound can be reformulated as:
\begin{equation}
\begin{split}
I(A,Q) &\geq \alpha[\sum _q P(q|a)\log P(q|a) + \sum _a P(a|q)\log P(a|q)] \\
    &\Longleftrightarrow {\sum_{<a,q>\in_{D_{aq}}}P(q|a)\log P(q|a) + P(a|q)\log P(a|q)}
\label{equation4}  
\end{split}
\end{equation}
where $\alpha$ denotes the sampling probability from a uniform distribution.

Through maximizing the above objective, we get two generation models $P(q|a;\theta_{aq})$ and $P(a|q;\theta_{qa})$.
These two models will indirectly supervise the subsequent bidword generation process as described in $L_{TRI}$.
In this way, bidword generation performance can be improved even if noisy data are used.

\subsection{Bidword as the Bridge between Query and Ads}
After constructing the generation model $A\Rightarrow Q$, it can indirectly supervise the generation process $A\Rightarrow B$ and $B\Rightarrow Q$ by taking $B$ as a bridge to connect $A$ and $Q$.
Given an <a, q> pair, our TRIDENT regards $B$ as an intermediate latent variable as follows:
\begin{equation}
\tilde{P}(q|a;\theta_{ab}, \theta_{bq})=\sum_b P(q|b;\theta_{bq})P(b|a;\theta_{ab})
\label{equation30}  
\end{equation}
Thus, $\tilde{P}(q|a;\theta_{ab},\theta_{bq})$ can be obtained by enumerating all possible $b$ that is relevant with both $a$ and $q$.

However, it is difficult to enumerate all possible candidates, as the search space is exponential to the size of vocabulary and the length of $b$ is unknown.
Instead, we leverage two approaches to compute $\tilde{P}(q|a;\theta_{ab},\theta_{bq})$: one is calculating an average vector space to represent the intermediate bidword variable~\cite{kovcisky2016semantic} (denoted as TRIDENT-A), the other is based on sampling method~\cite{kaelbling1996reinforcement} (denoted as TRIDENT-S).
% If we sample some sequences, the results will heavily depend on a carefully designed sampling strategy~\cite{kaelbling1996reinforcement}.
% Inspired by~\cite{kovcisky2016semantic}, we compute a sequence of expected word embedding for $k$ to represent the latent space.

Specifically, TRIDENT-A calculates a sequence of expected word embedding to represent the intermediate bidword variable.
Given an input $a$, the expected word embedding at each decoding step means the weighted average vector of all possible words:
\begin{equation}
\tilde{b_j}=\sum _{w\in V}P(w|w_{<j},a;\theta_{ab})Emb(w)
\label{equation31}
\end{equation}
where $j=1, 2, ...,  T_b$ and $T_b$ is a given length of $b$.
$V$ is the vocabulary and we enumerate each possible word $w$ in $V$.
$Emb(w)$ is the word embedding of $w$, and $P(w|w_{<j},a;\theta_{ab})$ is the weight which is the prediction probability for $w$ at $j$-th step.
Thus, a sequence $\tilde{b}=\{\tilde{b_1}, \tilde{b_2}, ..., \tilde{b_{T_b}}\}$ is obtained, which will be subsequently taken as input by model $B\Rightarrow Q$ to predict $q$:
\begin{equation}
\tilde{P}(q|a;\theta_{ab},\theta_{bq})=P(q|\tilde{b};\theta_{bq})
\label{equation32}
\end{equation}

Alternatively, TRIDENT-S computes $\tilde{P}(q|a;\theta_{ab},\theta_{bq})$ through a sampling method as follows:
\begin{equation}
\tilde{P}(q|a;\theta_{ab}, \theta_{bq})=\sum_{b\in C} P(q|b;\theta_{bq})P(b|a;\theta_{ab})
\label{equation33}
\end{equation}
$C$ is the candidate set, each of which is sampled from $P(b|a;\theta_{ab})$.
The sample size is set to $5$ in our experiments.
To make the loss differentiable, Gumbel-Softmax~\cite{jang2016categorical} is used here.

Likewise, $\tilde{P}(a|q;\theta_{qb},\theta_{ba})$ can be calculated in the same way.
% Susequently, $\tilde{p}(q|a;\theta_{ab}, \theta_{bq})$ and $\tilde{p}(a|q;\theta_{qb},\theta_{ba})$ will be supervised by $p(q|a; \theta_{aq})$ and $p(a|q; \theta_{qa})$ as in Equation\ref{equation555}.

\subsection{Training Details}
Overall, we first train the models $P(a|q)$ and $P(q|a)$ between $A$ and $Q$ based on a lower bound of their mutual information.
Then taking $B$ as a bridge to connect $A$ and $Q$, $\tilde{P}(a|q)$ and $\tilde{P}(q|a)$ can be computed through another four generation models ($P(a|b)$, $P(b|q)$, $P(q|b)$ and $P(b|a)$).
Subsequently, the cross-entropy between $P$ and $\tilde{P}$ is minimized, and through this way, $P(b|a)$ and $P(b|q)$ can be indirectly supervised in the triangle.
Thus, the quality of generated bidwords can be improved.
The detailed training process is summarized in Algorithm\ref{alg::training}.

\begin{algorithm}[ht] 
\caption{Training TRIDENT} 
\label{alg::training} 
\begin{algorithmic}[1]
\REQUIRE
A high-quality dataset $D_{aq}=\{<a,q>_i\}_{i=i}^N$ for $A$ and $Q$;\\
A noisy dataset $D_{ab}=\{<a,b>_i\}_{i=1}^M$ for $A$ and $B$; \\
A noisy dataset $D_{qb}=\{<q,b>_i\}_{i=1}^L$ for $Q$ and $B$; \\
Learning rate $\eta$;
\ENSURE 
Parameters $\theta_{aq}$, $\theta_{qa}$, $\theta_{ab}$, $\theta_{ba}$, $\theta_{qb}$ and $\theta_{bq}$
\REPEAT
\STATE sample a batch $B_{aq}$ of <$a$, $q$> pairs from $D_{aq}$
\STATE compute the gradient of the lower bound of mutual information defined in Equation\ref{equation4}: $g=\bigtriangledown I(B_{aq};\theta_{aq}, \theta_{qa})$
\STATE update parameters $\theta_{aq}$ and $\theta_{qa}$: $\theta =\theta + \eta * g$
\UNTIL{convergence}
\REPEAT
\STATE sample a batch $B_{aq}$ of <$a$, $q$> pairs from $D_{aq}$, $B_{ab}$ of <$a$, $b$> pairs from $D_{ab}$ and $B_{qb}$ of <$q$, $b$> pairs from $D_{qb}$
\STATE compute the gradient on loss $L$ in Equation\ref{equation1}: \\$g=\bigtriangledown L(B_{aq}, B_{ab}, B_{qb};\theta_{ab}, \theta_{ba}, \theta_{qb}, \theta_{bq})$
\STATE update parameters $\theta_{ab}$, $\theta_{ba}$, $\theta_{qb}$, $\theta_{bq}$: $\theta =\theta - \eta * g$
\UNTIL{convergence}
\end{algorithmic}
\end{algorithm}

\subsection{Constrained Beam Search}
\citet{li2016simple} find that most responses in the $N$-best candidates produced by the traditional beam search are similar. 
To solve this problem, we propose a constrained beam search to foster diversity in the generated bidwords.
We force the head words of $N$-candidates should be different, and then the model continues to generate a response by a greedy decoding strategy after such head words are determined.
Through such a simple method, our model can generate the $N$-best candidates with more diversity, which have great potential to bring extra revenues for sponsored search.

\section{Experiment}
\label{sec:experiment}
In this section, we conduct extensive experiments to show the performance of the proposed TRIDENT in generating reasonable bidwords for sponsored search.

\begin{table}[!t]
\begin{center}
\begin{tabular}{l|l|c|c|c}
\hline
\hline
\multicolumn{2}{c|}{Dataset} & Size & Vocabulary & Average Length \\
\hline
\multirow{2}{*}{(A., Q.)} & A. & $12,998,127$ & $119,806$ & $16.22$  \\
\cline{2-5}
& Q. & $14,634,482$ & $477,684$ & $4.67$ \\
\hline
\multirow{2}{*}{(A., B.)} & A. & $188,773$ & $28,113$ & $18.71$ \\
\cline{2-5}
& B. & $7,338,854$ & $188,629$ & $4.09$ \\
\hline
\multirow{2}{*}{(Q., B.)} & Q. & $11,913,539$ & $316,243$ & $4.62$ \\
\cline{2-5}
& B. & $2,540,295$ & $103,346$ & $2.86$\\
\hline
\hline
\end{tabular}
\end{center}
\caption{The detailed statistics of the datasets. A., Q. and B. denote Ad, query and bidword, respectively. }
\label{table1}
\end{table}

\begin{table*}[!t]
\begin{center}
\begin{tabular}{l|l|c|c|c|c|c}
\hline
\hline
\multicolumn{2}{l|}{\multirow{2}{*}{Models}} & \multicolumn{2}{|c|}{Relevance} & \multicolumn{3}{|c}{Diversity} \\
\cline{3-7}
\multicolumn{2}{c|}{} & Conv-KNRM & BLEU & Self-BLEU & distinct-3 & distinct-4  \\
\hline
\multirow{3}{*}{Match Models} & TF-IDF Method & $0.503$ & $0.063$ & $0.700$ & $0.119$ & $0.153$ \\
& Mean Pooling Method & $0.548$ & $0.098$ & $0.679$ & $0.147$ & $0.209$ \\
& Max Pooling Method & $0.548$ & $0.109$ & $0.673$ & $0.147$ & $0.209$ \\
\hline
\multirow{2}{*}{Neural Models} & Transformer\_base & $0.634$ & $0.133$ & $0.403$ & $0.166$ & $0.252$ \\
& MT-A2B & $0.657$ & $0.149$ & $0.483$ & $0.153$ & $0.236$ \\
\hline
\multirow{2}{*}{Our Models} & TRIDENT-S & $0.753$ & $0.151$ & $0.362$ & $0.168$ & $0.279$ \\
& TRIDENT-A & $\textbf{0.781}$ & $\textbf{0.185}$ & $\textbf{0.284}$ & $\textbf{0.218}$ & $\textbf{0.296}$ \\
\hline
\hline
\end{tabular}
\end{center}
\caption{Conv-KNRM, BLEU, Self-BLEU and distinct-3/4 results for advertising~(Ad) bidword generation.}
\label{table2}
\end{table*}

\begin{table*}[!t]
\begin{center}
\begin{tabular}{l|l|c|c|c|c|c}
\hline
\hline
\multicolumn{2}{l|}{\multirow{2}{*}{Models}} & \multicolumn{2}{|c|}{Relevance} & \multicolumn{3}{|c}{Diversity} \\
\cline{3-7}
\multicolumn{2}{c|}{} & Conv-KNRM & BLEU & Self-BLEU & distinct-3 & distinct-4  \\
\hline
\multirow{3}{*}{Match Models} & TF-IDF Method & $0.614$ & $0.104$ & $0.698$ & $0.208$ & $0.273$ \\
& Mean Pooling Method & $0.637$ & $0.135$ & $0.739$ & $0.215$ & $0.297$ \\
& Max Pooling Method & $0.651$ & $0.156$ & $0.729$ & $0.239$ & $0.336$ \\
\hline
\multirow{2}{*}{Neural Models} & Transformer\_base & $0.783$ & $0.186$ & $0.404$ & $0.258$ & $0.387$ \\
& Attn-Q2B & $0.801$ & $0.209$ & $0.517$ & $0.243$ & $0.358$ \\
\hline
\multirow{2}{*}{Our Models} & TRIDENT-S & $0.853$ & $0.205$ & $0.352$ & $0.286$ & $0.397$ \\
& TRIDENT-A & $\textbf{0.896}$ & $\textbf{0.239}$ & $\textbf{0.269}$ & $\textbf{0.381}$ & $\textbf{0.454}$\\
\hline
\hline
\end{tabular}
\end{center}
\caption{Conv-KNRM, BLEU, Self-BLEU and distinct-3/4 results for query bidword generation.}
\label{table3}
\end{table*}

\subsection{Dataset}

Since there are no off-the-shelf datasets for triangular bidword generation, we build the datasets by ourselves.
Three forms of data pairs are collected from a commercial search engine, which are <Ad, query> data, <Ad, bidword> data and <query, bidword> data, respectively.
Each item of <Ad, query> data represents a user click combined with a user issued query.
Each item of <Ad, bidword> data is an Ad and a related bidword purchased by the advertiser.
Each item of <query, bidword> data is a user issued query and a bidword purchased by the advertiser whose Ad is shown in the search page.
The detailed statistics of the three data pairs are shown in Table~\ref{table1}.
We randomly sample 5,000 pairs of each form as validation and test set.

\subsection{Experimental Settings}

We implement our proposed model with PyTorch\footnote{Available at https://pytorch.org/}.
Spercifically, our TRIDENT is trained using configuration \textit{transformer\_base}~\cite{vaswani2017attention}, which contains a $6$-layer encoder and $6$-layer decoder with $512$-dimensional hidden representations.
All the training data are first tokenized using the tokenizer provided by \citet{chang2008optimizing}, and then the words are split with a subword vocabulary learnt by BPE~\cite{sennrich2015neural}.
The size of the constrained beam search is set to $32$.
We tune the hyper-parameter $\lambda$ from $0.1$ to $1.0$ with step $0.1$, and find that $\lambda=0.6$ achieves the minimum perplexity on the validation set.

We apply Adam algorithm~\cite{kingma2014adam} as the optimizer with a linear warm-up over the first $4000$ steps and linear decay for later steps.
The batch size and learning rate are set to $32$ and $1e-4$, respectively.
The proposed model is trained for $100,000$ steps on $8$ Nvidia Telsa V100-32GB GPUs.

\subsection{Baseline Models}

We compare our proposed models against the following representative baselines, including matching based methods and neural network models:\\
(1) \textbf{Matching based methods} first convert an Ad/query into a feature vector, and then computes its relevance score with other Ads/queries by applying the cosine similarity function on their feature vectors. The bidword whose paired Ad/query achieves the highest similarity score is taken as the retrieved one. Three matching based methods are adopted:
\begin{itemize}
    \item \textbf{TF-IDF Method}~\cite{rajaraman2011mining} represents each Ad/query as a term tf-idf value vector.
    \item \textbf{Max Pooling Method}~\cite{wu2015max} represents each Ad/query as the max pooling of Glove word vectors~\cite{pennington2014glove}.
    \item \textbf{Mean Pooing Method}~\cite{10.1007/978-3-319-11331-9_81} represents each Ad/query as the mean pooling of Glove word vectors~\cite{pennington2014glove}.
\end{itemize}
(2) The following \textbf{Neural Network Models} are also taken as strong baselines:
\begin{itemize}
    \item \textbf{Transformer\_base}: We implement the \textit{transformer\_base} model as described in~\cite{vaswani2017attention} with the constrained beam search.
    \item \textbf{Attn-Q2B}: We reimplement Google's attention-based bidword generation model for query~\cite{du2018systems}.
    \item \textbf{MT-A2B}: The Ad-bidword generation model using multi-task learning~\cite{zhang2019improving} is also reimplemented.
\end{itemize}

% \begin{table*}[!t]
% \begin{center}
% \begin{tabular}{l|l|c|c|c|c|c|c}
% \hline
% \hline
% \multicolumn{2}{l}{\multirow{2}{*}{Models}} & \multicolumn{3}{|c|}{Beam Size=$8$} & \multicolumn{3}{|c}{Beam Size=$32$} \\
% \cline{3-8}
% \multicolumn{2}{c|}{}  & Relevance & Fluency & Diversity & Relevance & Fluency & Diversity \\
% \hline
% \multirow{2}{*}{Max Pooling Method} & A. to B. & $1.17$ & $1.48$ & $1.11$ & $1.23$ & $1.45$ & $\textbf{1.25}$ \\
% & Q. to B. & $1.27$ & $1.56$ & $1.04$ & $1.37$ & $1.47$ & $1.00$ \\
% \hline
% \multirow{2}{*}{Seq2Seq} & A. to B. & $1.25$ & $1.47$ & $1.12$ & $1.24$ & $1.54$ & $1.23$ \\
% & Q. to B. & $1.45$ & $1.57$ & $1.12$ & $1.43$ & $1.47$ & $1.11$ \\
% \hline
% \multirow{2}{*}{Neural Models} & MT-A2B & $1.36$ & $\textbf{1.52}$ & $1.02$ & $\textbf{1.39}$ & $\textbf{1.55}$ & $1.05$ \\
% & Attn-Q2B & $1.45$ & $\textbf{1.60}$ & $1.02$ & $1.43$ & $1.47$ & $1.05$  \\
% \hline
% \multirow{2}{*}{TRIDENT-A} & A. to B. & $\textbf{1.39}$ & $1.42$ & $\textbf{1.17}$ & $\textbf{1.39}$ & $1.52$ & $1.14$\\
% & Q. to B. & $\textbf{1.49}$ & $1.39$ & $\textbf{1.13}$ & $\textbf{1.44}$ & $\textbf{1.49}$ & $\textbf{1.15}$ \\
% \hline
% \hline
% \end{tabular}
% \end{center}
% \caption{The results of human evaluation. A., Q. and B. denote Ad, query and bidword respectively. {\em A. to B.} and {\em Q. to B.} represent bidwords generated from Ad and query, respectively.}
% \label{table4}
% \end{table*}

\subsection{Automatic Evaluation}
\subsubsection{Metrics}

The following automatic metrics are used to evaluate the performance of the proposed model:

\begin{itemize}

\item \textbf{Conv-KNRM}: Conv-KNRM~\cite{dai2018convolutional} models n-gram soft matches for ad-hoc search. Previous works have validated that Conv-KNRM can, to a large extent, capture the semantic-level similarity between queries and documents. Conv-KNRM here is trained on the <Ad, query> dataset, in which the click rate is taken as the similarity score.

\item \textbf{BLEU Score}: BLEU \cite{papineni2002bleu} is a popular metric that calculates the word-overlap score of the generated bidwords against gold-standard ones.

\item \textbf{Self-BLEU}: Self-BLEU is a metric to evaluate the diversity of the generated bidwords~\cite{zhu2018texygen}.
Regarding one bidword as hypothesis and the others as references, we calculate its BLEU score~\cite{papineni2002bleu}. 
Finally, Self-BLEU is the average BLEU score of all candidates. 

\item \textbf{Distinct}: Distinct-$1/2/3/4$ is the proportion of the distinct uni-grams/bi-grams/tri-grams/four-grams in all the generated tokens~\cite{li2016diversity}.
Here we report the distinct-3/4 results, which are more contrastive.
Distinct metrics can be used to evaluate the diversity of the generated bidwords.

\end{itemize}

% Above all, Conv-KNRM and BLEU score assess the relevance of the produced bidwords, while Self-BLEU and distinct-$3/4$ evaluate the diversity.

\begin{figure*} \centering    
\subfigure[Bidword generation results from query] {
 \label{fig:a}     
\includegraphics[width=0.65\columnwidth]{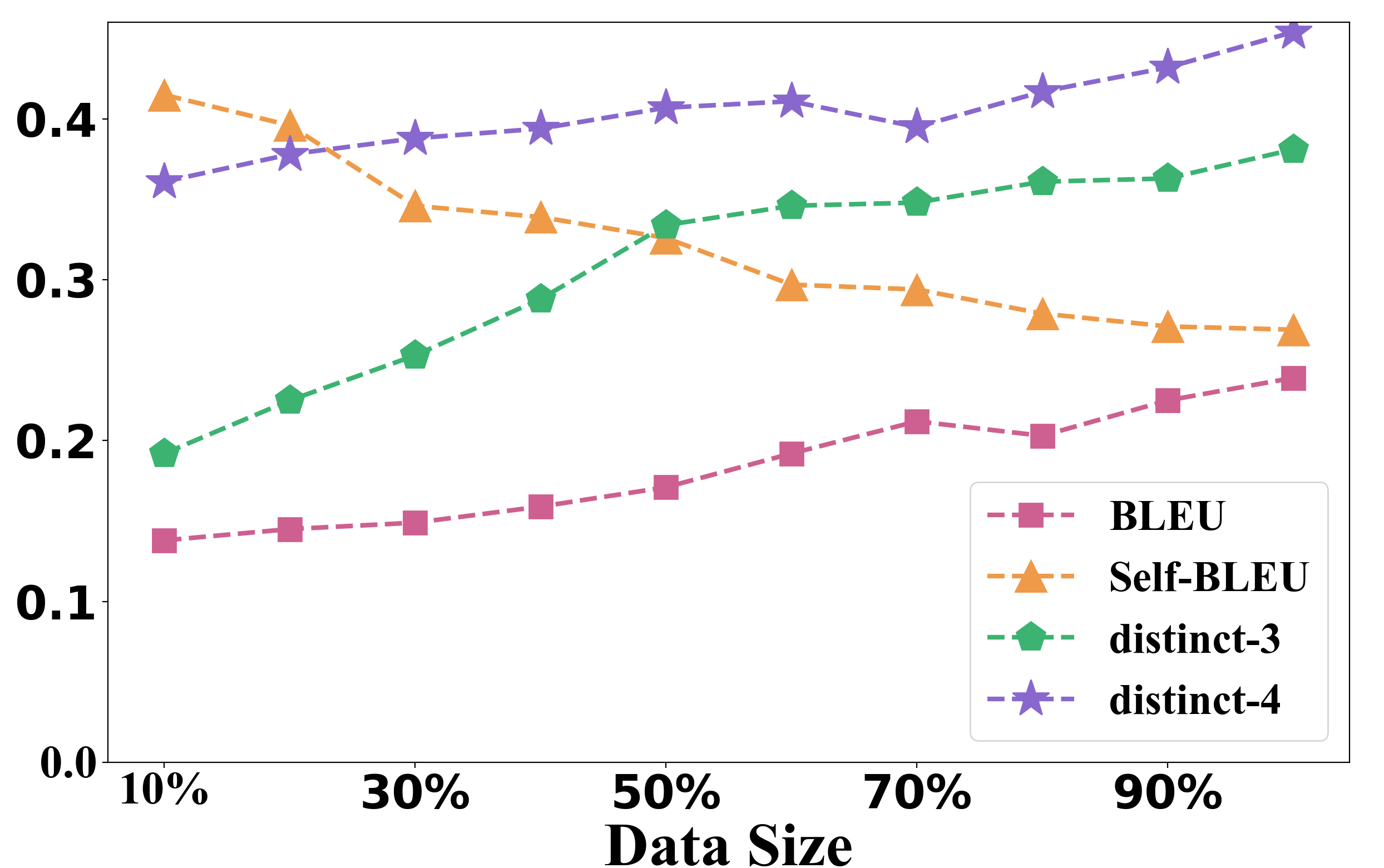}  
}     
\subfigure[Conv-KNRM results from Ad and query] { 
\label{fig:b}     
\includegraphics[width=0.65\columnwidth]{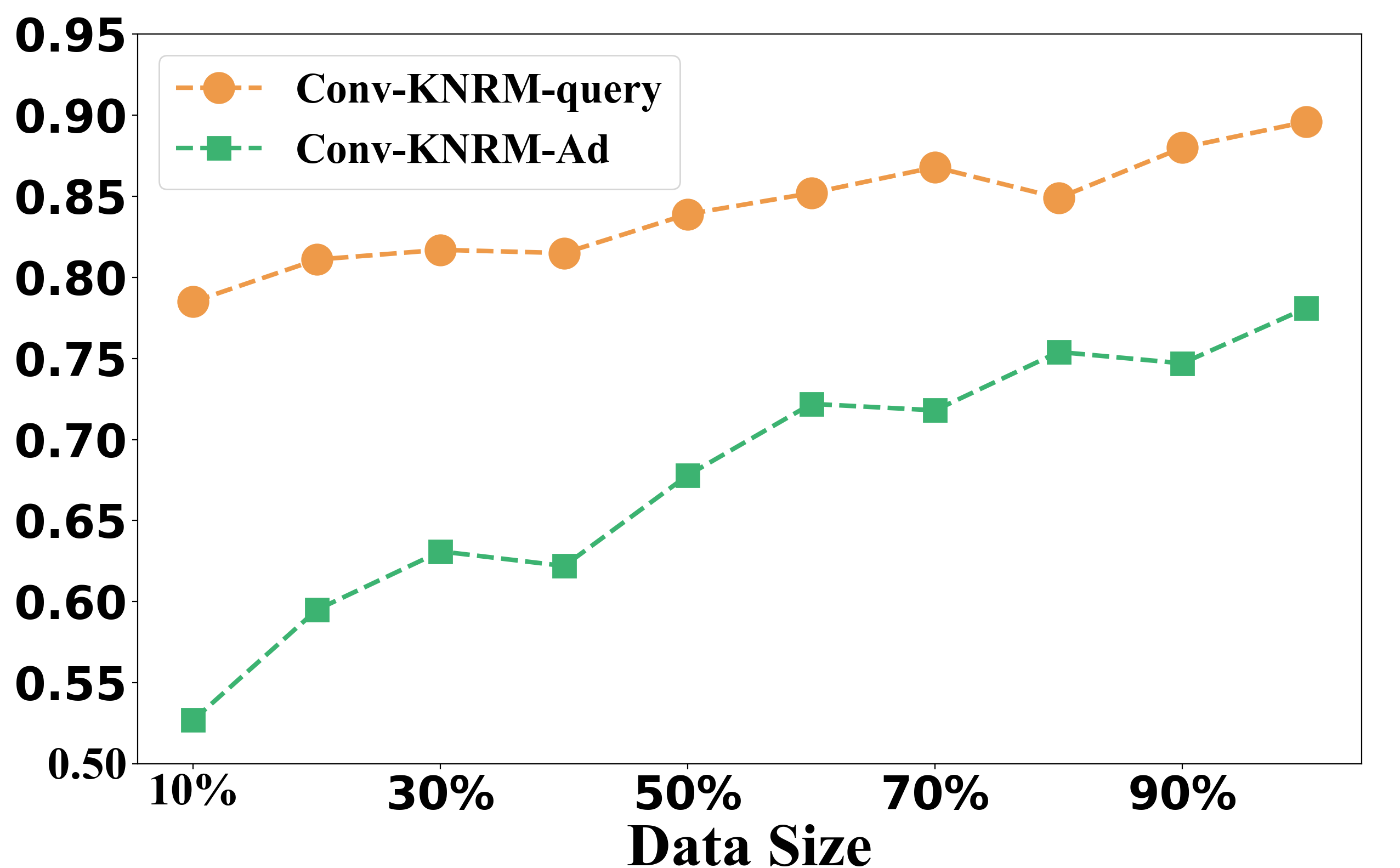}     
}    
\subfigure[Bidword generation results from Ad] { 
\label{fig:c}     
\includegraphics[width=0.65\columnwidth]{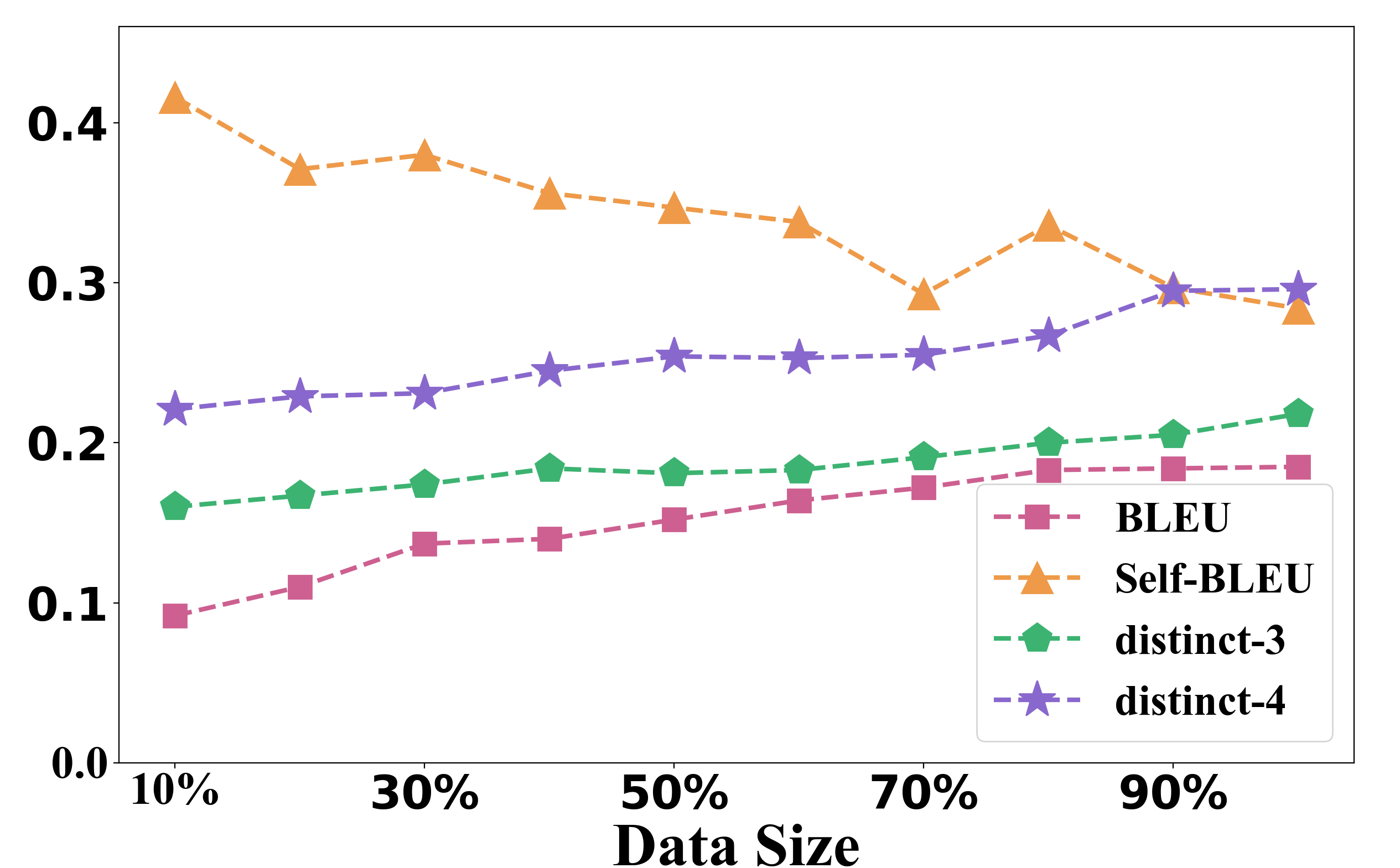}     
}   
\caption{ Performance curve of TRIDENT-A  with different size of <$A$, $Q$> data }     
\label{figure4}     
\end{figure*}

\begin{table}[!t]
\small
\begin{center}
\begin{tabular}{l|l|c|c|c}
\hline
\hline
\multicolumn{2}{l|}{Models} & Relevance & Diversity & Fluency \\
\hline
\multirow{2}{*}{Match Models} & A. to B. & $1.38$ & $1.80$ & $\textbf{1.97}$ \\
& Q. to B. & $1.29$ & $1.34$ & $\textbf{1.92}$ \\
\hline
\multirow{2}{*}{Transformer\_base} & A. to B. & $1.64$ & $1.36$ & $1.94$ \\
& Q. to B. & $1.64$ & $1.63$ & $1.72$ \\
\hline
\multirow{2}{*}{Neural Models} & MT-A2B & $1.42$ & $1.73$ & $1.96$ \\
& Attn-Q2B & $1.65$ & $1.25$ & $1.86$  \\
\hline
\multirow{2}{*}{TRIDENT-A} & A. to B. & $\textbf{1.76}$ & $\textbf{1.84}$ & $1.95$\\
& Q. to B. & $\textbf{1.70}$ & $\textbf{1.64}$ & $1.80$ \\
\hline
\hline
\end{tabular}
\end{center}
\caption{The results of human evaluation. A., Q. and B. denote Ad, query and bidword, respectively. {\em A. to B.} and {\em Q. to B.} represent bidwords generated from Ad and query, respectively.}
\label{table4}
\end{table}

\subsubsection{Results}

The bidword generation results from advertising and query are reported in Table~\ref{table2} and Table~\ref{table3}, respectively.
\textbf{As shown in the two Tables, our model (TRIDENT-A) outperforms the competitors in all cases.}

%相比另外两个neural models，看相关性；相比match models看多样性；最后对比两个trident
\textbf{Notably, the bidwords generated by our models are significantly better than baselines in both relevance and diversity whether from advertising or query}.
In Table~\ref{table2}, MT-A2B performs best among all baselines in relevance, while our proposed TRIDENT-A exceeds it on Conv-KNRM with a fairly significant margin of $0.124$ points.
Moreover, our TRIDENT-A is also superior to MT-A2B on BLEU score with a significant boost about $3.6$ points.
The similar results can be observed in the Table~\ref{table3}.
It demonstrates that our TRIDENT are capable of generating more relevant bidwords than other neural models.
The reason is that the triangular architecture built in TRIDENT can exploit the high-quality <Ad, query> data to improve the performance of the noisy data (<query, bidword> and <Ad, bidword> data).
Additionally, our TRIDENT-A and Transformer\_base significantly outperform other models in diversity.
The Self-BLEU scores achieved by our TRIDENT-A/S and Transformer\_base are much lower than other models in both Table~\ref{table2} and Table~\ref{table3}.
This makes sense because generative models can generate novel bidwords that do not exist in the training corpus and the proposed constrained beam search can further promote the diversity.
Besides, we also observe that TRIDENT-A performs better than TRIDENT-S in all cases, demonstrating that the expected word embedding method works better than sampling method.

% \textbf{The proposed constrained beam search is able to promote diversity} in the generated candidates.
% % and generally diversified keywords are preferred by users.
% As shown in Table\ref{table2}, TRIDENT-A-CBS/TRIDENT-S-CBS/Seq2Seq-CBS obviously performed better than TRIDENT-A-BS/TRIDENT-S-BS/Seq2Seq-BS with an average increment of $9.53\%$, $9.72\%$ and $13.3\%$ respectively in diversity (including distinct-1/2 and Self-BLEU).
% The similar results in Table\ref{table3} again verifies the effect of the constrained beam search.
%后面做分离实验

% \textbf{Our TRIDENT can generate more relevant bidwords when the beam size gets larger}, which will better satisfy a commercial search engine.
% When the beam size is increased from $20$ to $40$, our model (TRIDENT-A-BS/CBS) obtained an average reduction of $1.51\%$ on the embedding score, while the baseline (Seq2Seq-BS/CBS) got one of $1.59\%$.
% It demonstrates that the triangular architecture is helpful to generate more relevant candidates through leveraging another large-scale high-quality <Ad, query> data.

\begin{table*}[!t]
\begin{center}
\begin{tabular}{l|l|c|c|c|c|c}
\hline
\hline
\multicolumn{2}{l|}{\multirow{2}{*}{Models}} & \multicolumn{2}{|c|}{Relevance} & \multicolumn{3}{|c}{Diversity} \\
\cline{3-7}
\multicolumn{2}{c|}{} & Conv-KNRM & BLEU & Self-BLEU & distinct-3 & distinct-4  \\
\hline
\multirow{2}{*}{Transformer\_base} & Beam Search & $0.800$ & $0.175$ & $0.639$ & $0.135$ & $0.211$ \\
& Constrained Beam Search & $0.783$ & $0.186$ & $0.404$ & $0.258$ & $0.387$ \\
\hline
\multirow{2}{*}{TRIDENT-A} & Beam Search & $0.784$ & $0.228$ & $0.518$ & $0.270$ & $0.346$ \\
& Constrained Beam Search & $\textbf{0.896}$ & $\textbf{0.239}$ & $\textbf{0.269}$ & $\textbf{0.381}$ & $\textbf{0.454}$\\
\hline
\hline
\end{tabular}
\end{center}
\caption{Results of constrained beam search against beam search.}
\label{table5}
\end{table*}

\subsection{Human Evaluation}
\subsubsection{Evaluation Settings}

We conduct a human evaluation to better understand the quality of the bidwords generated by our triangular architecture. 
In this way, we can decide if these bidwords are suitable to be recommended to advertisers.

Specifically, $50$ Ads and $50$ queries are first randomly sampled from the test set of <Ad, bidword> and <query, bidword> data. 
All neural models and max pooling method, which performs best among all matching based methods, are selected as baselines to compare with our proposed TRIDENT-A.
For each of the Ads and queries, all models generate the corresponding bidwords with beam size $32$.
Later the pairs of <Ad, top-$32$ bidwords> and <query, top-$32$ bidwords> are presented to five human annotators with order disrupted.
Then they evaluate the produced bidwords at relevance, fluency and diversity levels all by $3$-scale rating ($0$, $1$, $2$).
A higher score means a better performance.
Relevance assesses whether the bidwords are coherent and meaningful for an Ad or query.
Fluency tests the general grammar correctness.
Diversity decides if the top-$32$ bidwords are diverse.

Agreements to measure inter-rater consistency among the annotators are calculated with the Fleiss's kappa~\cite{fleiss1973equivalence}.
As a result, the Fleiss's kappa for relevance, fluency and diversity is $0.857$, $0.912$ and $0.803$ respectively, all of which indicate "substantial agreement".

\subsubsection{Results}
The human evaluation results are reported in Table~\ref{table4}.
\textbf{It shows that TRIDENT-A achieves the highest scores in both relevance and diversity (2-tailed sign test, p − value < $0.05$).}

As we can see, TRIDENT-A outperforms MT-A2B/Attn-Q2B in relevance with a significant margin of $0.34$/$0.05$ points. 
Besides, TRIDENT-A also achieves higher scores than Transformer\_base in relevance.
These observations indicate that our proposed TRIDENT-A is capable of generating more relevant bidwords than other neural models.
Our interpretation is that some extra information can be leveraged through the triangular architecture, which can provide an indirect supervision for bidword generation.
Moreover, the diversity scores of TRIDENT-A are also higher than baselines, indicating that a better bidword generation model can be obtained through the triangular framework and the proposed constrained beam search can further promote the diversity.
Notably, the max pooling method performs best in fluency. 
It is easy to understand as matching based methods directly select bidwords from the corpus, which generally have no grammar errors.
% This demonstrates that the proposed constrained beam search can truly promote diversity in the generated candidates due to the fact that different head words could boost more diverse sentences~\cite{li2016simple}. 

\begin{table}[!t]
\begin{center}
\setlength{\tabcolsep}{3mm}{
\begin{tabular}{l|c|c|c}
\hline
\hline
Model & Precision & Recall & F$1$-score \\
\hline
Transformer\_base & $0.156$ & $0.075$ & $0.101$ \\
\hline
TRIDENT-A & $\textbf{0.214}$ & $\textbf{0.115}$ & $\textbf{0.150}$\\
\hline
\hline
\end{tabular}}
\end{center}
\caption{Evaluation on online real data.}
\label{table6}
\end{table}

\subsection{Additional Analysis}
We further investigate the influence of the used <Ad, query> data size and the proposed constrained beam search.

\subsubsection{Effect of the <Ad, Query> Data Size}
In order to investigate the influence of the high-quality <Ad, query> data size, we test the performance of our TRIDENT-A with different <Ad, query> data size.

The performance curve with different <Ad, query> data size is shown in Figure~\ref{figure4}.
Specifically, the TRIDENT-A achieves a higher Conv-KNRM and BLEU score when trained on more <Ad, query> data.
It demonstrates that our TRIDENT can generate more relevant bidwords when more <Ad, query> data are used.
This makes sense because more data could provide stronger supervision on the bidword generation process, leading to more meaningful bidwords.
Moreover, the Self-BLEU of our model gets lower with the increase of <Ad, query> data size. 
One explanation is that more <Ad, query> data can offer guide information to more training samples of bidwords.

% 一张图for self-bleu and distinct-2
% 一张图for bleu and Conv-KNRM

\subsubsection{Effect of the Constrained Beam Search}
To clearly show the effect of the proposed constrained beam search, we examine the model  performance using constrained beam search against the original beam search~\cite{reddy1977speech}.
The results are reported in Table~\ref{table5}.
As we can see, the Self-BLEU of constrained beam search are much lower than that of original beam search for both Transformer\_base and TRIDENT-A.
It demonstrates that the proposed constrained beam search can promote diversity in the generated candidates due to the fact that different head words could boost more diverse sentences~\cite{li2016diversity}

\begin{table*}[!t]
\small
\begin{center}
\begin{tabular}{l|l|l|l|l}
\hline
\hline
\multicolumn{1}{c|}{Ad/Query} & \multicolumn{1}{|c|}{Max Pooling Method} & \multicolumn{1}{|c|}{Transformer\_base} & \multicolumn{1}{|c}{MT-A2B/Attn-Q2B} & \multicolumn{1}{|c}{TRIDENT-A}\\
\hline
\multirow{11}{*}{} & 今日股票行情  & 股票行情分析 & 股票软件 & 今日股市行情 \\
 & (Today's stock market) & (Stock market analysis) & (Stock software) & (Today's stock market)  \\
\cline{2-5}
& 股价 & 个股分析 & 股价 & 石油股价 \\
股市行情怎样？ & (Share price) & (Stock analysis) & (Share price) & (Oil stock price)  \\
\cline{2-5}
立马下载APP了解！ & 精股 & 走势图 & 股市行情 & 炒股 \\
How is the stock market? & (Featured stocks) & (Tendency photos) & (Stock market) & (Trading stocks)  \\
\cline{2-5}
Download the APP right & 分析 & 科技股 & 大盘 & 低价股 \\
now to follow it! & (Analysis) & (Technology stocks) & (Market) & (Low-priced stocks)  \\
\cline{2-5}
\textbf{}
 & 消息 & 利好消息 & 交流软件 & 科技股 \\
 & (News) & (Good news)  & (Communication software) & (Technology stocks) \\
\cline{2-5}
\hline
\multirow{15}{*}{} & 手镯翡翠  & 翡翠手镯价格 & 玉镯 & 极品翡翠手镯 \\
& (Jade bracelet)  & (Price of Jade bracelet)  &  (Jade bracelet)  & (The best Jade bracelet)  \\
\cline{2-5}
& 玉镯 & 鉴定 & 冰种翡翠手镯  & 珠宝翡翠 \\
翡翠手镯 & (Jade bracelet) & (Identification)   & (Ice Jadeite bracelet) & (Jewelry and  jade)  \\
\cline{2-5}
(Jade bracelet) & 玉手镯 & 买翡翠手镯 & 镯子 & 天然翡翠手镯 \\
 & (Jade bracelet) & (Buy jade bracelet)  & (Bracelet)  & (Natural jade bracelet) \\
\cline{2-5}
& 购买翡翠手镯 & 天然翡翠手镯 & 玉石 & 帝王绿翡翠 \\
& (Buy jade bracelet)  & (Natural jade bracelet) & (Jade) & (Emperor green jade) \\
\cline{2-5}
& 天然翡翠手镯  & 好的翡翠手镯 & 直播 & 玉手镯 \\
& (Natural jade bracelet)  & (Good jade bracelet)  & (Live broadcast)  & (Jade bracelet)\\
\cline{2-5}
\hline
\hline
\end{tabular}
\end{center}
\caption{Case study for the proposed TRIDENT. The top-$5$ bidwords are generated for each example. Sentence in the parentheses is the corresponding translation.}
\label{table7}
\end{table*}

\subsection{Evaluation on Online Real Data}
We additionally simulate an online experiment to examine whether the generated bidwords can be correctly recommended to relevant products in a real world e-commerce search engine.
Specifically, we collect three pairs of data among A, Q and B from January $1$ to July $15$, 2020, of which the data before and after July $1$ are split into training and test set, respectively.
Taking <Q, B> pair as an example, we compute the precision, recall and F$1$-score between the generated bidwords and the golden ones for each query.
The results are reported in Table~\ref{table6}, which shows the F$1$-score of our model is higher than that of Transformer\_base.
Therefore, we can draw the conclusion that the bidwords generated by our model are more popular with advertisers, which will help to recall more related products.

\subsection{Case Study}
To gain an insight on how well the bidwords are generated through the proposed triangular architecture, we provide some examples.
We randomly sample one Ad and one query as distinct sources, and collect the generated top-$5$ bidwords from all models.

The results are reported in Table~\ref{table7}, from which we can see that the bidwords generated by our TRIDENT are more relevant with the given Ad or query than other models.
For example, given a query ``翡翠手镯 (Jade bracelet)'', other models may generate some irrelevant bidwords, such as ``鉴定 (Identification)'' and ``直播 (Live broadcast)'', while those generated by our TRIDENT-A are all pointful and have different meanings.
Above all, the proposed TRIDENT is capable of generating better bidwords in both relevance and diversity.

\section{Related Work}
\label{sec:related}
\subsection{Bidword Generation}
A lot of models have been proposed for bidword generation, since bidwords take an important part in sponsored search.
Previous works are mostly retrieval based methods.
\citeauthor{joshi2006keyword}~\cite{joshi2006keyword} construct a graph model to generate relevant bidwords from queries based on their similarity scores.
% Abhishek and Hosanagar et al. 
\citeauthor{abhishek2007keyword}~\cite{abhishek2007keyword} further establish a kernel function to improve the calculation of similarity scores.
% Later Fuxman at al. 
\citeauthor{fuxman2008using}~\cite{fuxman2008using} propose to random walk with absorbing states for bidword generation.
\citeauthor{chen2008advertising}~\cite{chen2008advertising} propose a bidword suggestion method that exploits the semantic knowledge among concept hierarchy to find bidwords through the shared concepts.

More recently, deep learning based methods have been applied for bidword recommendation with various neural network structures.
% Grbovic et al. 
\citeauthor{grbovic2015context}~\cite{grbovic2015context} apply three language models to learn distributed query representations to promote query expansion.
% \citeauthor{zhai2016deepintent}~\cite{zhai2016deepintent} provide an attention network on top of the RNN to assign attention scores to different words, and they have verified that this attention network helps RNNs learn better vector representations for ads and queries. 
% Lian et al. 
\citeauthor{lian2019end}~\cite{lian2019end} directly employ the neural machine translation framework to generate bidwords from queries.
% Zhou et al. 
\citeauthor{zhou2019domain}~\cite{zhou2019domain} use a reinforcement learning algorithm to generate domain constrained bidwords.
\citeauthor{du2018systems}~\cite{du2018systems} predict the CTR between query and bidword using an attention model.
\citeauthor{zhang2019improving}~\cite{zhang2019improving} apply a multi-task learning framework to imporve the semantic similarity between advertising and bidword.

All the above methods generate bidwords directly from Ad or query (say Ad bidword suggestion or query expansion), while our model exploits a triangular architecture to generate bidwords from both Ad and query.

\subsection{Multilingual Neural Machine Translation}
% Multilingual training of Neural Machine Translation (NMT) becomes more and more popular these years as it has led to impressive accuracy improvements on low resource languages \cite{wang2019multilingual}.
Multilingual training of Neural Machine Translation (NMT) has brought impressive accuracy improvements on low resource languages~\cite{lin2020pre}.
% Dong et al. 
\citeauthor{dong2015multi}~\cite{dong2015multi} extend the single NMT to a multi-task learning framework that shares source language representation and separates the modeling of different target language translation.
% Zoph et al. 
\citeauthor{zoph2016transfer}~\cite{zoph2016transfer} propose a transfer learning method to improve the performance of low-resource language pairs.
\citeauthor{firat2016multi}~\cite{firat2016multi} enable a single neural translation model for each language pair and all the language pairs share a single attention mechanism.
A TA-NMT model is exploited in~\cite{ren2018triangular} to improve the performance of the low resource pair by constructing a triangular architecture.

To the best of our knowledge, this is the first work exploiting a triangular framework to generate more relevant and diverse bidwords with the help of the high-quality <Ad, query> data.

\subsection{Neural Text Generation}
Recently generative neural networks have been proven to be quite successful in structured prediction tasks such as machine translation~\cite{zheng2019mirror}, 
% document summarization~\cite{emnlp-rash:15}, 
bidword generation~\cite{lian2019end} and query expansion~\cite{lee2018rare}.

% Sutskever et al. 
\citeauthor{sutskever2014sequence}~\cite{sutskever2014sequence} propose a sequence-to-sequence (Seq2Seq) framework for machine translation.
%where a source sentence is first transformed to a vector representation by an encoder and then this representation is fed into a separate decoder to generate the target sentence.
% Bahdanau et al. 
\citeauthor{bahdanau2014neural}~\cite{bahdanau2014neural} then extend the Seq2Seq framework with an attention mechanism.
Seq2Seq framework has also been used in many natural language generation tasks, such as abstractive summarization~\cite{rush2015neural} and dialog generation~\cite{shi2020dispersed,song2019generating}.
However, neural generative models employing the Seq2Seq framework tend to produce generic and meaningless outputs~\cite{li2016simple}.
% \citeauthor{shang2015neural}~\cite{shang2015neural} then apply this Seq2Seq framework to dialogue systems.
To address this problem, \citeauthor{miao2019cgmh} propose CGMH~\cite{miao2019cgmh}, a constrained generation method using Metropolis-Hastings sampling to improve sentence generation quality.
Further, a novel objective function based on maximizing mutual information is proposed by~\citeauthor{li2016diversity}~\cite{li2016diversity} to generate more meaningful dialogue responses.
\citeauthor{song2020improving} employs adversarial training to improve both the quality and diversity of generated texts~\cite{song2020improving}. 
% Serban et al. 
% \citeauthor{serban2017hierarchical}~\cite{serban2017hierarchical} introduce a hierarchical encoder-decoder framework with latent variables to boost the diversity of the responses.
% An emotional dialogue system called EmoDS is further proposed in \cite{song2019generating} to promote emotional and diverse responses.
% A NEXUS network connecting both the preceding dialogue history and the following conversations is further proposed in \cite{shen2018nexus} to promote long and pointful responses.
Recently, the Transformer encoder-decoder framework~\cite{vaswani2017attention} is also employed in text generation models~\cite{vlasov2019dialogue} to boost coherence.

\section{Conclusion}
\label{sec:conclusion}
Sponsored search auction is also called bidword auction, which is an indispensable part of commercial search engine.
Generating relevant and diverse bidwords for search queries and advertisements are crucial and can bring great revenues in sponsored search.
However, directly employing the Seq2Seq model in bidword generation does not yield satisfactory results, due to the fact that the Seq2Seq model is data-driven and heavily relies on the training data, while the training data of <query/advertisement, bidword> may be noisy because of the \textit{keyword bidding problem}.
In this paper, we propose a Triangular Bidword Generation Model (TRIDENT), which exploits the high-quality <query, advertisement> data as a supervision signal to indirectly guide the bidword generation process through a triangle training framework.
Experimental results show that our proposed \model\ can generate relevant and diverse bidwords for both queries and advertisements, which are more popular with advertisers.

\bibliography{main.bbl}
\bibliographystyle{ACM-Reference-Format}

\end{CJK*}
\end{document}